\documentclass[conference]{IEEEtran}
\usepackage{breakurl}
\usepackage{xurl}
\usepackage[noadjust]{cite}
\usepackage{graphicx}
\usepackage{algpseudocode}
\usepackage{algorithm}
\usepackage{tabularx}
\usepackage{caption}
\usepackage{subcaption}
\usepackage{makecell}
\usepackage[table]{xcolor}
\usepackage{array}
\newcolumntype{P}[1]{>{\centering\arraybackslash}p{#1}}
\newcolumntype{M}[1]{>{\centering\arraybackslash}m{#1}}
\usepackage[edges]{forest}
\usepackage{subcaption}
\usepackage{tikz}
\usepackage{hyperref}
\usepackage{pgfplots}
\usepackage{array,multirow,graphicx}
\usetikzlibrary{patterns}
\usepackage{rotating}
\usepackage{amsmath}

\usepackage{algorithm, algpseudocode}
\usepackage{xcolor}
\usepackage{gensymb}
\usepackage{placeins}
\usetikzlibrary{decorations.pathreplacing,positioning}
\usepackage{graphicx} 
\usepackage{tabularx} 
\usepackage[noadjust]{cite}
\DeclareRobustCommand*{\IEEEauthorrefmark}[1]{%
  \raisebox{0pt}[0pt][0pt]{\textsuperscript{\footnotesize\ensuremath{#1}}}}
  \usepackage[misc]{ifsym}
\newcommand{\corrAuthor}{$^{\textrm{\Letter}}$}
\ifCLASSINFOpdf
\else
\fi

\DeclareRobustCommand*{\IEEEauthorrefmark}[1]{%
  \raisebox{0pt}[0pt][0pt]{\textsuperscript{\footnotesize\ensuremath{#1}}}}
  \usepackage[misc]{ifsym}
\begin{document}

\title{Determining HEDP Foams' Quality\\with Multi-View Deep Learning Classification}

\author{\IEEEauthorblockN{Nadav Schneider\IEEEauthorrefmark{1,2},
Matan Rusanovsky\IEEEauthorrefmark{3,4},
Raz Gvishi\IEEEauthorrefmark{5} and
Gal Oren\IEEEauthorrefmark{3,6\hspace{0.1cm}$\corrAuthor$}}\\
\IEEEauthorblockA{\IEEEauthorrefmark{1}Technology Division, Soreq Nuclear Research Center - Yavne, Israel}
\IEEEauthorblockA{\IEEEauthorrefmark{2}Israel Atomic Energy Commission}
\IEEEauthorblockA{\IEEEauthorrefmark{3}Scientific Computing Center, Nuclear Research Center – Negev, Israel}
\IEEEauthorblockA{\IEEEauthorrefmark{4}Department of Physics, Nuclear Research Center – Negev, Israel}
\IEEEauthorblockA{\IEEEauthorrefmark{5}Applied Physics Division, Soreq Nuclear Research Center - Yavne, Israel}
\IEEEauthorblockA{\IEEEauthorrefmark{6}Department of Computer Science, Technion – Israel Institute of Technology, Israel}
{\tt\small nadavsc@soreq.gov.il, matanr@nrcn.org.il, rgvishi@soreq.gov.il galoren@cs.technion.ac.il}}

\IEEEtitleabstractindextext{%
\begin{abstract}
High energy density physics (HEDP) experiments commonly involve a dynamic wave-front propagating inside a low-density foam. This effect affects its density and hence, its transparency. A common problem in foam production is the creation of defective foams. Accurate information on their dimension and homogeneity is required to classify the foams' quality. Therefore, those parameters are being characterized using a 3D-measuring laser confocal microscope. For each foam, five images are taken: two 2D images representing the top and bottom surface foam planes and three images of side cross-sections from 3D scannings. An expert has to do the complicated, harsh, and exhausting work of manually classifying the foam's quality through the image set and only then determine whether the foam can be used in experiments or not. Currently, quality has two binary levels of normal vs. defective. At the same time, experts are commonly required to classify a sub-class of normal-defective, i.e., foams that are defective but might be sufficient for the needed experiment. This sub-class is problematic due to inconclusive judgment that is primarily intuitive. In this work, we present a novel state-of-the-art multi-view deep learning classification model that mimics the physicist's perspective by automatically determining the foams' quality classification and thus aids the expert. Our model achieved 86\% accuracy on upper and lower surface foam planes and 82\% on the entire set, suggesting interesting heuristics to the problem. A significant added value in this work is the ability to regress the foam quality instead of binary deduction and even explain the decision visually. 
\\The source code used in this work, as well as other relevant sources, are available at: \textcolor{blue}{\url{https://github.com/Scientific-Computing-Lab-NRCN/Multi-View-Foams.git}}.\\
\end{abstract}
\begin{IEEEkeywords}
HEDP, Low-Density Foams, Aerogel, Multi-View Classification, Deep Learning, LIME.
\end{IEEEkeywords}}


\maketitle

\IEEEdisplaynontitleabstractindextext

%
\IEEEpeerreviewmaketitle

\section{Introduction}
\subsection{HEDP Foams and Aerogels}
\begin{figure}[!hb]
    \subfloat{{\includegraphics[height=5cm]{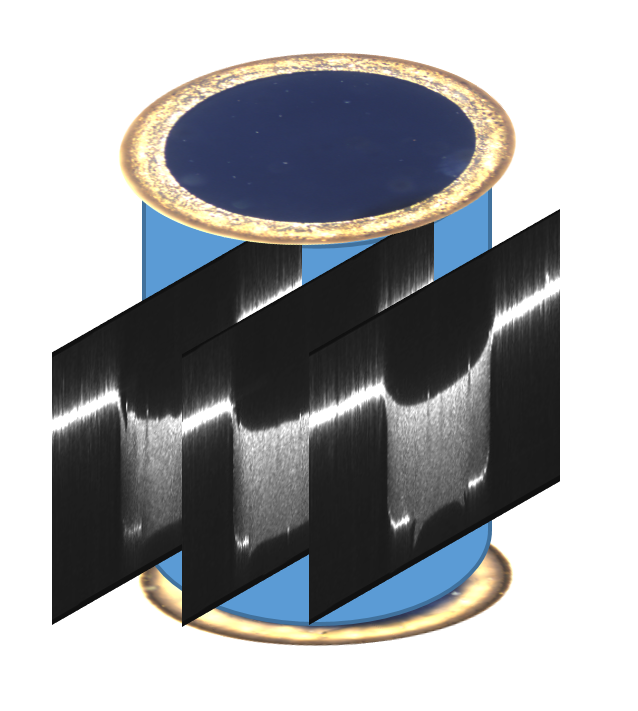}}}
    \qquad
    \subfloat{{\includegraphics[height=5cm]{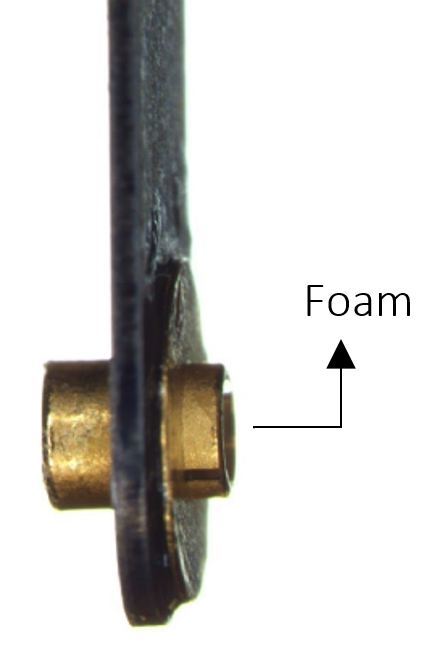}}}%
    \caption{Illustration of the acquired microscope's images (left) from a given foam (right).}%
    \label{fig:images/foam}%
\end{figure}

\begin{figure}[!ht]
    \centering
    \includegraphics[height=1.7cm]{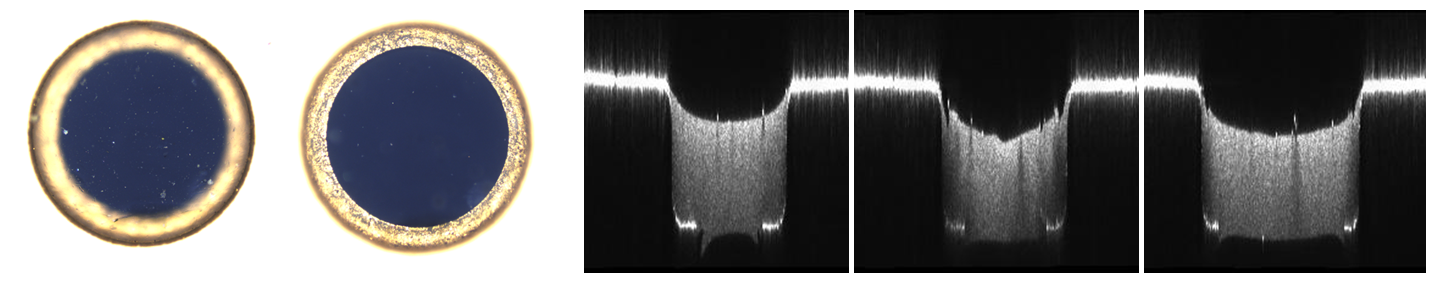}
    \caption{Five images of an original example from the data set. From left to right: top and bottom views and three profiles.}
    \label{fig:original_examples}
\end{figure}

High energy density physics (HEDP) experiments \cite{ref1, ref2, ref3, ref4, harel2020complete} commonly involve a dynamic wave-front propagating inside a low-density foam. This effect affects its density and hence, its transparency. The analysis of the experimental measurements required accurate information on the dimension and homogeny of the foam. Therefore, a dimension and homogeny characterization of the foam using a 3D-measuring laser confocal microscope is needed. For each foam, five images were taken: two 2D images representing the upper surface foam plane (henceforth, 'top' and 'bottom' images) and three other images of side cross-sections from 3D-scanning images (henceforth, 'profiles' images) (Fig. \ref{fig:images/foam}, \ref{fig:original_examples}).

The foam used in the HEDP experiments is commonly aerogel. Aerogels are a large family of materials, generally defined as extremely low-density solids (more than 90\% porosity, less than 200mg/cm$^3$ density) \cite{ref7, ref8}. Aerogels can be composed of metals or dielectric materials and can be produced in pure, hybrid, and doped forms. For these reasons, aerogels can exhibit a diverse range of properties (chemical and physical) that could also be tailored for a specific application. Aerogels are signified by their unique properties, especially low density and high specific surface area (commonly above $\sim$600m$^2$/g), resulting in an excellent insulating material for heat, electric and acoustic. Therefore, their applications range from thermal insulators \cite{ref9, ref10, ref11, ref12}, acoustic insulation \cite{ref13, ref14}, catalyst supports, electrode materials and fuel cell \cite{ref15, ref16, ref17}, random lasers matrices \cite{ref18}, space micrometeorites collectors \cite{ref19}, bio-medical \cite{ref20}, drug-delivery \cite{ref21, ref22, ref23}, cosmetic, lightweight magnetic actuator \cite{ref24}, Cherenkov radiators \cite{ref25, ref26} and HEDP measurements \cite{ref2, ref4}.

\begin{figure*}[!ht]
\captionsetup[subfigure]{labelformat=empty}
\begin{subfigure}{.375\textwidth}
\centering
\includegraphics[height=6cm]{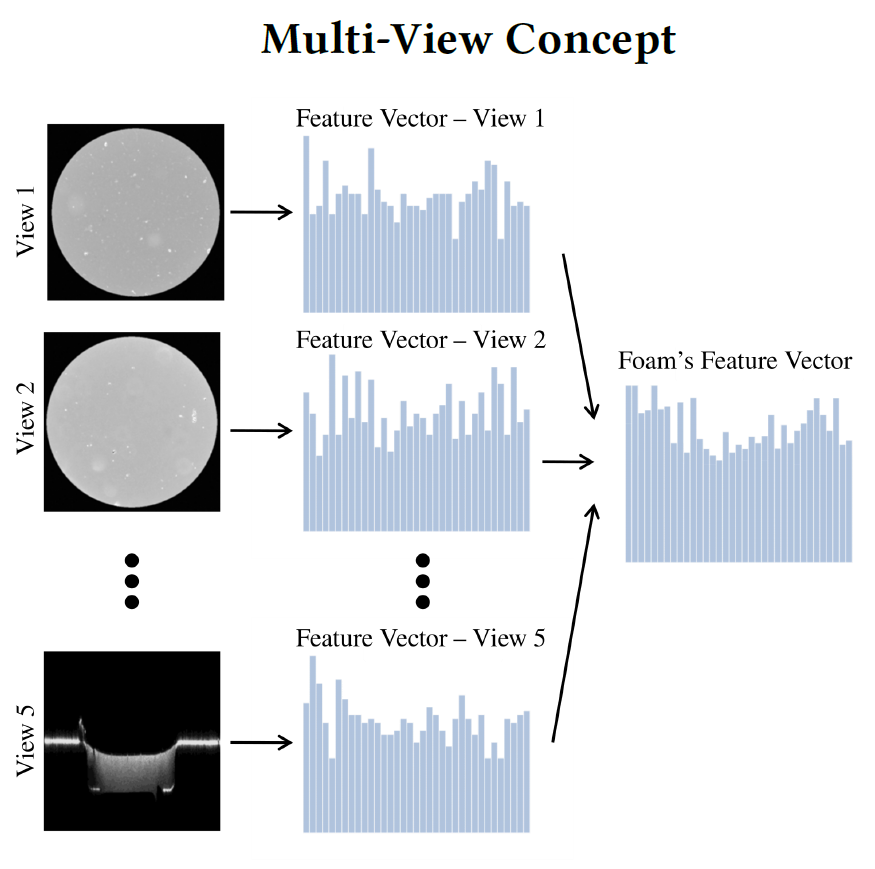}
\caption{(A)}
\end{subfigure}
\begin{subfigure}{.35\textwidth}
\centering
\includegraphics[height=6cm]{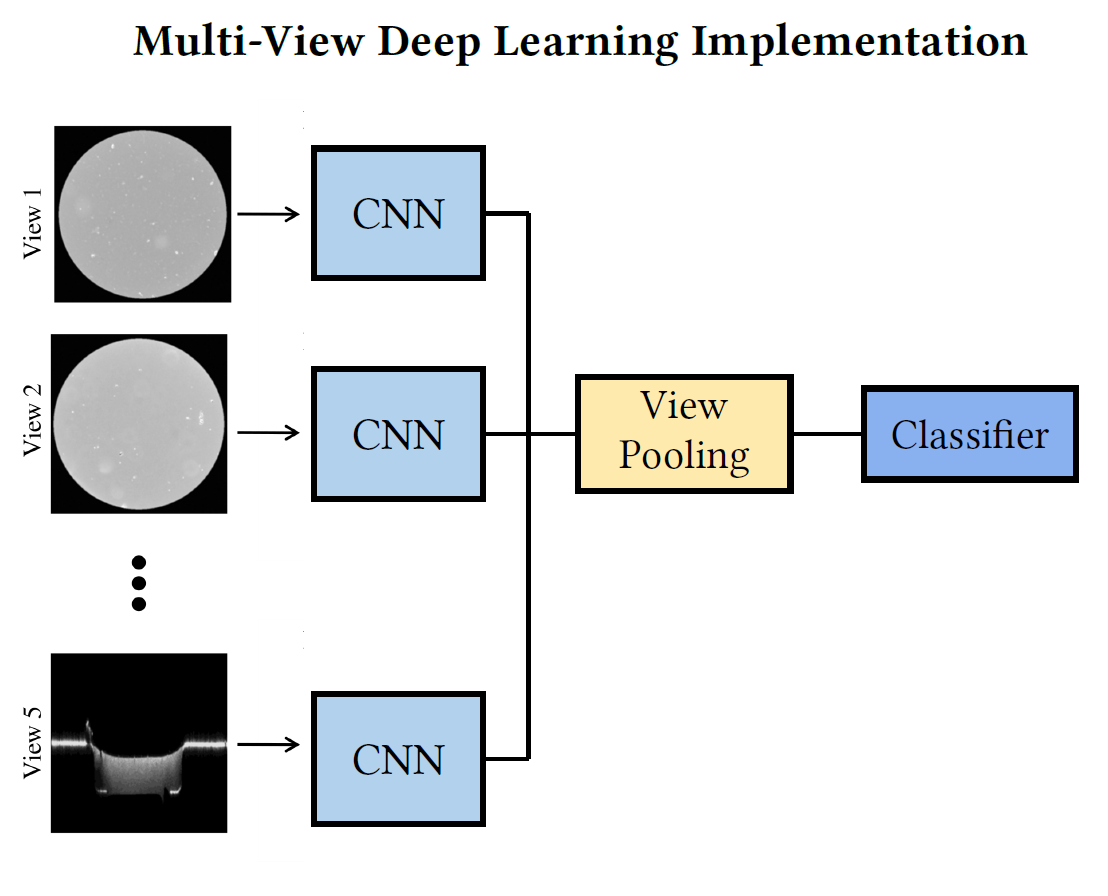}
\caption{(B)}
\end{subfigure}
\begin{subfigure}{.35\textwidth}
\centering
\includegraphics[height=5.9cm]{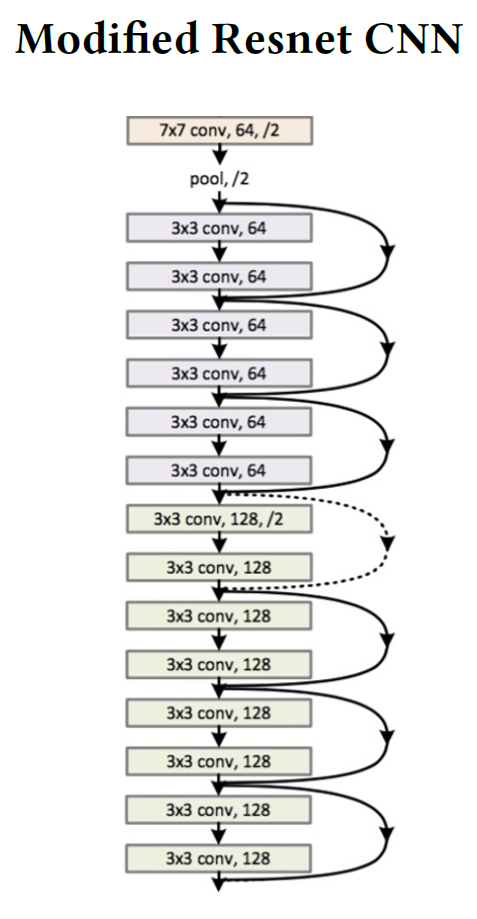}
\caption{(C)}
\end{subfigure}
\caption{(A) The multi-view concept, (B) the chosen multi-view deep learning implementation approach, and (C) represents the modified Resnet in each CNN block that was intentionally reduced to prevent overfit due to a lack of data.}
\label{fig:multi_view_schema}
\end{figure*}


\subsection{Determinning HEDP Foams' Quality: Current Status}
Aerogels are commonly prepared from Si-alkoxide precursors through sol-gel chemistry \cite{ref27, ref28, ref29, ref30}. The gel formed is made of a few nanometers thick of fragile walls in a random structure surrounded by meso-sized pores. To dry these delicate materials, supercritical drying is necessary. Transferring the pore fluid to the supercritical phase makes it possible to vent it out with no capillary forces \cite{ref31}. Thus, supercritical drying is essential to achieve dry material without a collapse of the fine porous structure. A significant limitation of dry aerogels is mechanical fragility and tending to suffer from cracks \cite{ref32, ref33, ref34}. Therefore, it is required to characterize each sample for its quality as a foam. The expert searches for characteristics such as scratches, dirt, and dark stains in the top and bottom views images, implying a deep hole inside the foam. As a complementary, the profile images help confirm or refute the initial assumptions. This complicated, harsh, and exhausting work of manually classifying the foams' quality through the image set is mandatory to decide whether said foams can be used in experiments (such as HEDP). Currently, quality has binary levels of normal vs. defective. Experts are also commonly required to classify a sub-class of normal-defective, i.e., foams that are defective in some way but might be sufficient for a given experiment. This sub-class is problematic due to inconclusive judgment that is primarily intuitive. Thus, the need to devise a new, precise, explainable, consistent, and objective classification of said foams' quality is essential.

\subsection{Suggested New Approach: Multi-View Classification}
In order to mimic the physicist's intuitive perspective and automatically determine the foams' quality even in borderline cases, there is a need to devise either an intelligent or even a learned model, which will be able to self-conclude the desired classification \cite{rusanovsky2022end, rusanovsky2020mlography}. Moreover, as the image set, in this case, is diverse -- both in 3D perspective (top and bottom views) and in acquisition technology (optics vs. x-ray), standard classification methodologies that do not consider perspective and domain heterogeneity will fail. Thus, the recently novel multi-view approach \cite{xu2013survey, sun2013survey, zhao2017multi} was chosen to solve this problem correctly. 

Multi-view classification models are designed for cases when the final decision depends on different features in different images \cite{seeland2021multi}. Multi-view approach suggests looking over an object from several points of view (both spatially and visually), which is technically achieved by extracting unique features of the object from all the viewpoints, concatenating them into one vector, and making the decision on it. The idea has two implementation approaches: (1) Classical feature extractions and machine learning algorithms \cite{nie2017multi, houthuys2018multi}. (2) Deep multi-view adjusted convolutional neural networks \cite{kan2016multi, seeland2021multi}. In this work, we examined both of the methods as our data set is comparably small for initial learning.



\vspace{0.1cm}
\subsubsection{Features Extraction and Machine Learning}
Before the age of deep learning, there was a need to "manually" extract features from given images \cite{kumar2014detailed}. This technique aims to numerically distinguish and represent unique features of the image \cite{kumar2014detailed}. Several feature extractor algorithms were devised over the years for different tasks. Some of the famous ones are SIFT (Scale Invariant Feature Transform) and SURF (Speeded-Up Robust Features) \cite{juan2009comparison}. The main focus of these feature extractors is extracting unique features such as scale, illumination, rotation, distortion, etc. Most feature extractors collect key points of interest in the image and produce a descriptor for each. The descriptors are usually numerical vectors with constant size.

In the multi-view approach, the descriptors are collected from the different images and grouped into centroids \cite{nie2017multi, houthuys2018multi} using an unsupervised model such as $K$-Means \cite{hartigan1979algorithm}. These centroids are like a "vocabulary of visual words" \cite{multiweb}. The next stage is finding visual words for each image and a vector creation that counts the number of features belonging to each centroid of visual words. The result is a $(1, K)$ properties vector representing an image, and using this process on all the images in the data set yields a $(N, K)$ matrix while $N$ represents the number of images and $K$ represents the vocabulary size. Afterward, classification is done on the matrix using a machine learning algorithm such as logistic regression, supported vector machine (SVM) \cite{houthuys2018multi}, etc. Finally, solution adaptation into multi-view is done by creating a feature vector for all images in one example, using a feature extractor, and counting the visual words for all images (Fig. \ref{fig:multi_view_schema} A). 

Nevertheless, in this work, no learning or understanding could be achieved using this method, suggesting either not enough data was collected for the data set or no sufficient learning was made. As such, there was a need for a comprehensive deep-learning approach.

\vspace{0.1cm}
\subsubsection{Deep Multi-View Adjusted Convolutional Neural Networks}
Following the success of convolutional neural networks in computer vision problems \cite{khan2018guide}, there was an attempt to adjust these networks to multi-view classification \cite{su2015multi}. One approach is to take the feature map of every view and stack them. For example, if every feature map is in size $K \cdot K \cdot C$, a union of the five maps is executed along the channels' dimensions, and one matrix of $K \cdot K \cdot 5C$ of feature maps is created. This approach assumes there is an order of orientation between the different views. The other approach is pooling the feature maps (view pooling), which is the chosen approach for the given problem (although orientation exists, the images do not naturally precept a single object from different angles). This approach unites all the views' features without assuming an order (Fig. \ref{fig:multi_view_schema} B).

In computer vision, when there are merely several examples similar to the given problem, as in our case, transfer learning, which uses prior knowledge accumulated from another model trained on another data set, is usually used \cite{weiss2016survey}. ResNet is the selected pre-trained model for the given problem \cite{he2016deep}, trained on the ImageNet data set \cite{russakovsky2015imagenet}, which contains millions of high-resolution labeled images and 22 thousand different classes. This model is designed for object detection tasks using feature extraction, and with an addition of neuron layers at the end of the network, one can finetune for a given task.

\section{Data set Creation}
While many data sets are created with strict rules and repeatable methods for an adequate examination, our data set was created solely for an expert's classification purpose. In the absence of reproducible settings for the 3D laser microscope and an arranged process of collecting the data, a considerable amount of pre-processing is required to fit the data set into a machine learning model. Furthermore, since only experts can label these examples, the labeled data is rare, and there is no real option to reach extensive labeled data. Consequently, our goal in this work was to develop an end-to-end model with the constraint of a severe shortage in data. Iteratively, we added batches of new labeled images when they were ready while designing the model as a few-shot model. This iterative process results in a data set with 95 labeled examples, each consisting of 5 images, i.e., 475 images in total.

We notice that the normal-defective labels are given when an example does not have an obvious decision for its quality. Without significant model learning, which can be achieved using lots of normal and defective examples, one can not expect to classify samples even an expert could not. Therefore, we refer to the normal-defective labels as defective labels under the strict assumption of "if there is a doubt, there is no doubt." This assumption is necessary for the given problem because we can not afford to ignore examples due to their tiny amount. However, the severe assumption above is not always correct, leading to a trade-off between the measured metrics; accuracy and AUC (Section \ref{sec:conclusion}). Thus, we check our model on different learning configurations (Fig. \ref{fig:pyramid}).

\section{Pre-learning Process}
An initial running of the multi-view deep convolutional neural networks on the data set has been done, and a quickly evolved overfit been observed. There are two common reasons for overfitting: a few training examples and an over-complicated model \cite{kadam2018review}. Therefore, solutions such as data augmentation, reduction, and pre-processing (Table \ref{table:pre_process}, Fig. \ref{fig:augmentation_examples}) are done to overcome the few physics-guided examples issue while preserving the original and relevant properties for decision.

\begin{table}[ht!]
    \begin{center}
     \begin{tabular} { M{0.25cm} | M{2cm} | M{2cm} | }
        \centering
       & Original & Pre-Proccesed \\  
      \hline
        \rotatebox[origin=c]{90}{\centering {\thead{Top/Bottom}}}  & 
          \includegraphics[height=2cm,width=2cm]{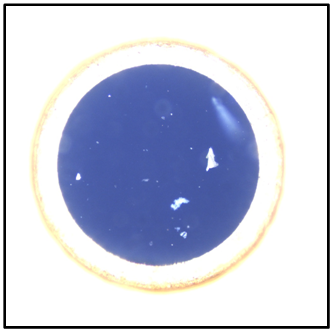} &
          \includegraphics[height=2cm,width=2cm]{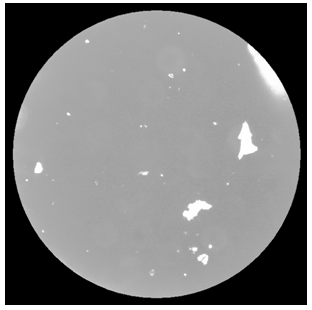}
        \\ \hline
        \rotatebox[origin=c]{90}{\centering {\thead{Profiles}}} & 
          \includegraphics[height=2cm,width=2cm]{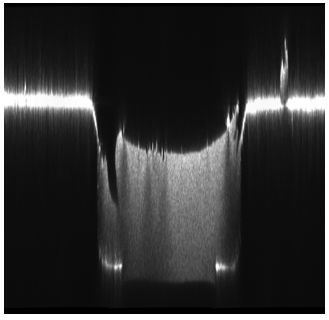} &
          \includegraphics[height=2.477cm,width=1.33cm]{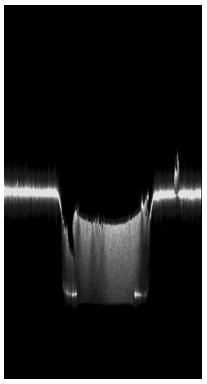}
          \\ \hline 
          
      \end{tabular}
      \end{center}
      \caption{Data set images before and after pre-processing.}
      \label{table:pre_process}
\end{table}

\subsection{Data Augmentation}
\begin{figure}[!ht]
    \centering
    \includegraphics[height=1.8cm]{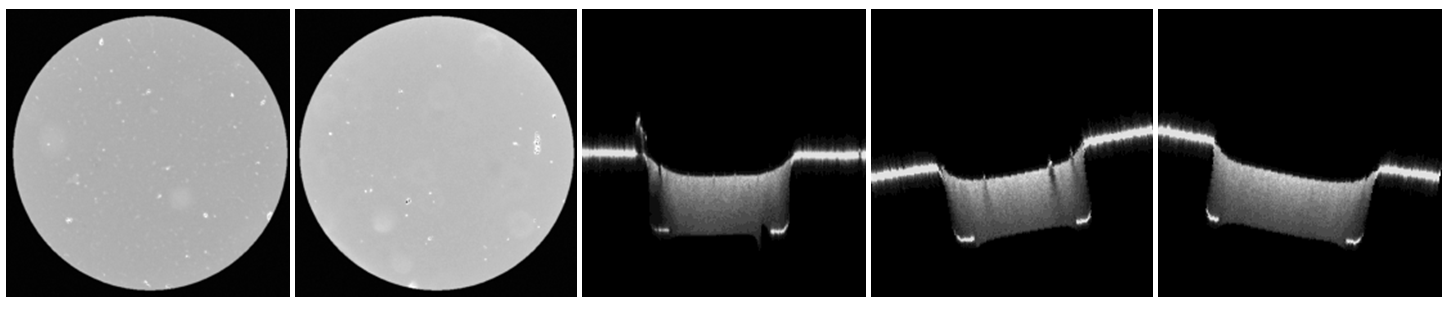}
    \caption{Five images of an original example from the data set after pre-process and data augmentation stages. From left to right: top and bottom views and three profiles.}
    \label{fig:augmentation_examples}
\end{figure}

Orientation conservation is easily destroyed when creating new examples with augmentation. Nonetheless, since training deep learning models require lots of examples, an augmentation approach has been attempted. Classical augmentations such as brightness, contrast adjustments, and rotations contribute to the robustness and generalization of the model and, therefore, decrease overfit. Several rotations permutations have been chosen: -10\degree{} to 10\degree{} with a 5\degree{} jump for the profiles and 0\degree{} to 270\degree{} with a 90\degree{} jump for the top and bottom views images. This wide range is possible due to the symmetry of the images (circles). In addition, gaussian noise, brightness, and contrast adjustments have been performed (Fig. \ref{fig:augmentation_examples}).

\subsection{Data Reduction and Pre-Processing}
\begin{figure}[!ht]
    \centering
    \includegraphics[height=3.0cm]{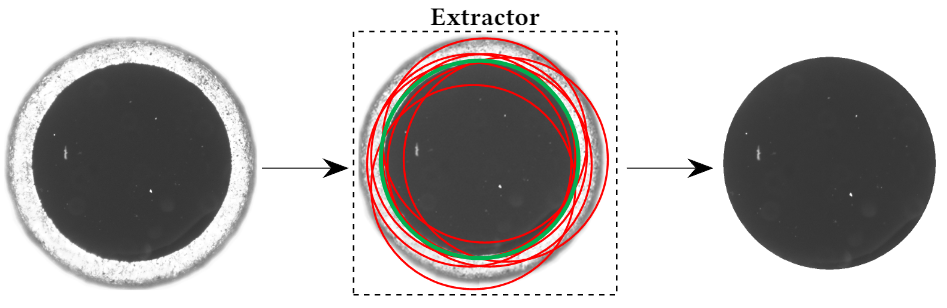}
    \caption{Automatic circle extractor algorithm. Choosing the circle with minimal bright pixels and maximal dark pixels.}
    \label{fig:circle_extractor}
\end{figure}

Pre-process helps focus the model on relevant features for classification. An automatic physics-guided pre-processing tool has been developed to create a proper data reduction that preserves the original and relevant properties of the foam with the following actions:
\begin{enumerate}
    \item \textit{RGB to grayscale conversion}: classification of the foams is independent of their color; therefore, the conversion prevents the model from learning irrelevant features.
    \item \textit{Intensity quantization}: a division of the pixels' values (which range from 0 to 255) to 10 bins. In many images, the black background is not an absolute black but a collection of values near zero. The model might search for a reason around these pixels' variance, and bins division prevents it.
    \item \textit{Circle extraction}: only the central circle is relevant for classification. Extracting this circle\footnote{In order to focus the model on relevant features, an automatic algorithm for circle extraction in the top and bottom views images is required. Canny edge detection \cite{canny1986computational} using erosion and dilation is not enough to effectively mask the circles. Hence, a designated algorithm has been developed. The algorithm uses permutations of circles in different radii and locations around the center to find the circle with minimal bright pixels and maximal dark pixels (Fig. \ref{fig:circle_extractor}). The algorithm assumes that the background will always be brighter than the circle. Pixel's definition of bright or dark is changeable.} and masking the ring and other outliers helps the model focus on relevant features (Fig. \ref{fig:circle_extractor}).
    \item \textit{Circles bounding}: minimizing the black background solves the problem of variance in the circles' position and conserves homogeneity between the images.
    \item \textit{Profiles centering and padding}: padding black pixels for the background of profiles if necessary and centering the profiles. These operations conserve homogeneity and constant size between the profiles. Further manual cropping has been done to keep only the relevant scope.
\end{enumerate}

\section{The Learning Process and its Results}
\label{sec:conclusion}


\begin{table*}
\begin{minipage}{\columnwidth}
    \centering
 \begin{tabular}{||c c c c c||} 
 \hline
  ACCURACY & One-view & One-view-ND & Multi-view & Multi-view-ND \\ [0.5ex] 
 \hline\hline
 Top & 72 & 78 & -- & -- \\ 
 Bottom & 79 & 84 & -- & -- \\
 Top-Bottom & 77 & 73 & \textbf{86} & 78 \\
 Profiles & -- & -- & 79 & 73 \\
 Full Group & -- & -- & 82 & 78 \\ [1ex] 
 \hline
 \end{tabular}
    \caption{Accuracy for each purposed model.}
    \label{table:accuracy}
\end{minipage}\hfill 
\begin{minipage}{\columnwidth}
    \centering
\begin{tabular}{||c c c c c||} 
 \hline
  AUC & One-view & One-view-ND & Multi-view & Multi-view-ND \\ [0.5ex] 
 \hline\hline
 Top & 74 & 76 & -- & -- \\
 Bottom & 76 & 74 & -- & -- \\
 Top-Bottom & 69 & 78 & 71 & 64 \\
 Profiles & -- & -- & 77 & 81 \\
 Full Group & -- & -- & \textbf{84} & 75 \\ [1ex] 
 \hline
 \end{tabular}
    \caption{AUC for each purposed model.}
    \label{table:auc}
  \end{minipage}
\end{table*}

\begin{table*}
    \begin{center}
     \begin{tabular} { M{0.5cm} | M{4cm} | M{4cm} | M{4cm} | M{4cm} |}

       & Loss +ND & Accuracy +ND & Loss -ND & Accuracy -ND  \\  
      \hline
        \rotatebox[origin=c]{90}{\centering {OV Top}}  & 
          \includegraphics[height=3.0cm,width=4cm]{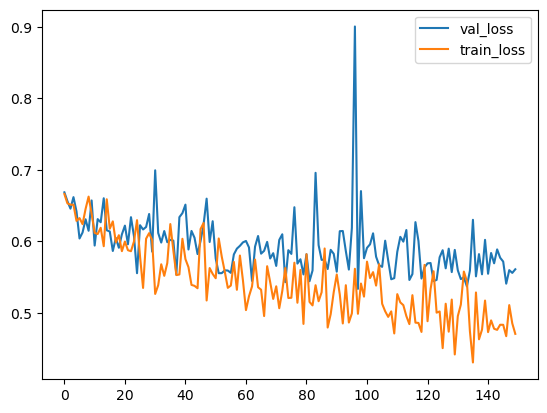} &
          \includegraphics[height=3.0cm,width=4cm]{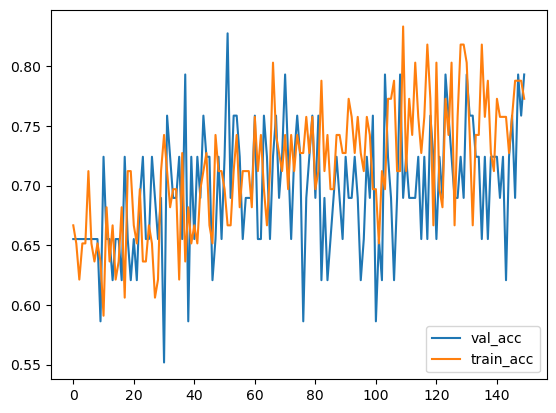} &
          \includegraphics[height=3.0cm,width=4cm]{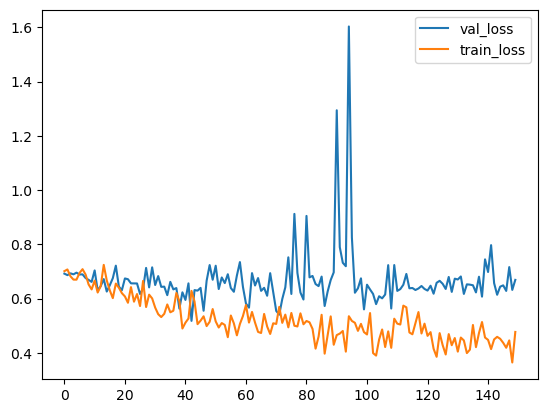} &
          \includegraphics[height=3.0cm,width=4cm]{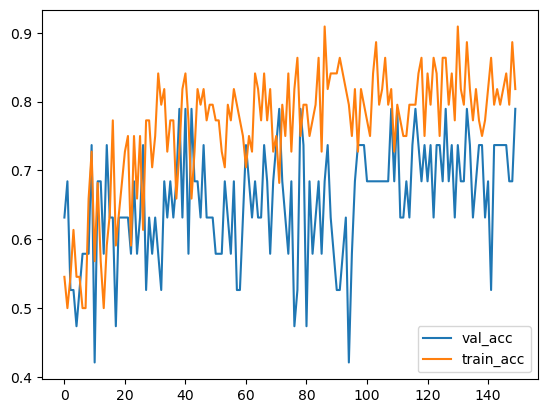}
        \\ \hline

        \rotatebox[origin=c]{90}{\centering {OV Bottom}} & 
          \includegraphics[height=3.0cm,width=4cm]{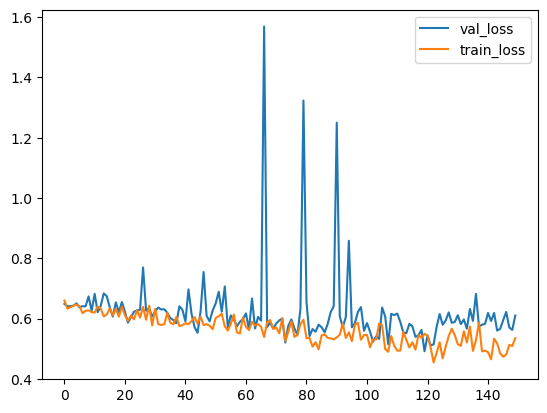} &
          \includegraphics[height=3.0cm,width=4cm]{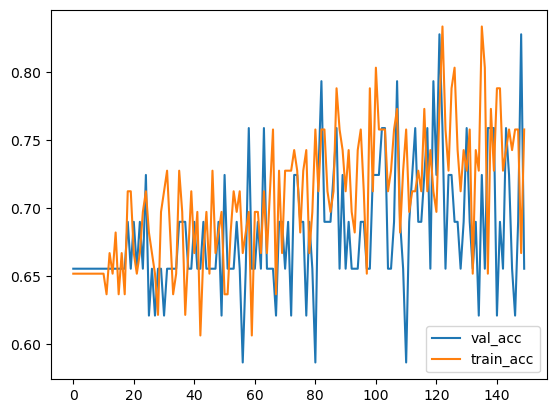} &
          \includegraphics[height=3.0cm,width=4cm]{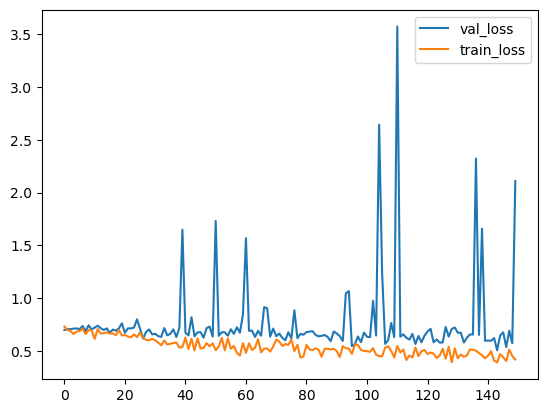} &
          \includegraphics[height=3.0cm,width=4cm]{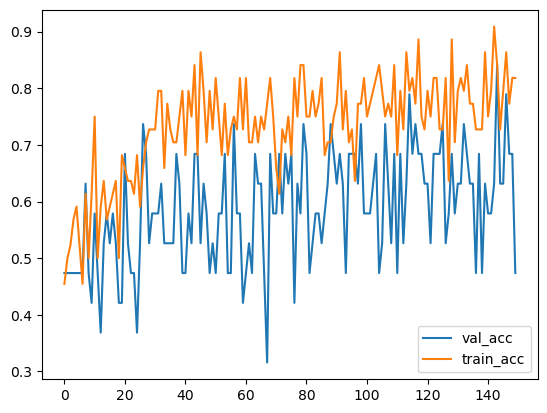}
          \\ \hline 

        \rotatebox[origin=c]{90}{\centering {OV Top Bottom}} &
          \includegraphics[height=3.0cm,width=4cm]{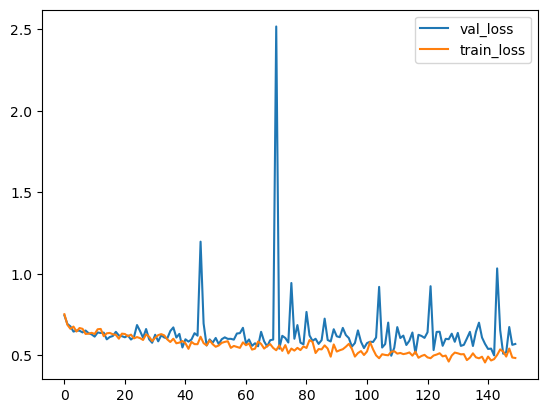} &
          \includegraphics[height=3.0cm,width=4cm]{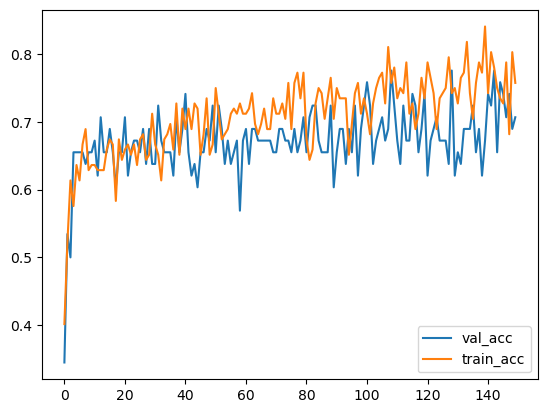} &
          \includegraphics[height=3.0cm,width=4cm]{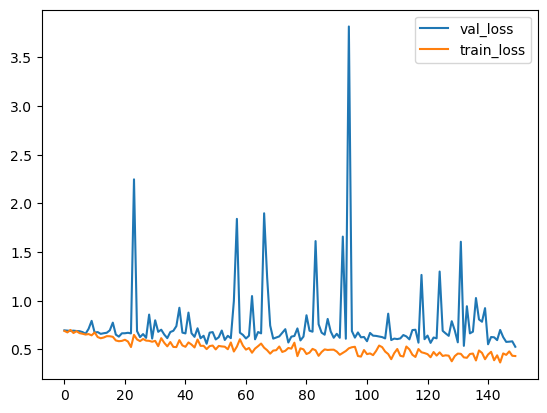} &
          \includegraphics[height=3.0cm,width=4cm]{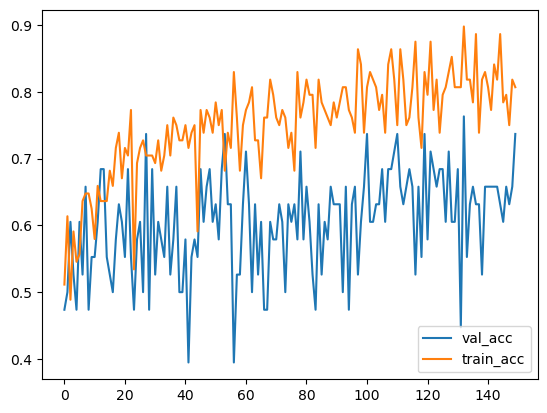}
        \\ \hline
        
        \rotatebox[origin=c]{90}{\centering {MV Top Bottom}} &
          \cellcolor[HTML]{D5DBDB} \includegraphics[height=3.0cm,width=4cm]{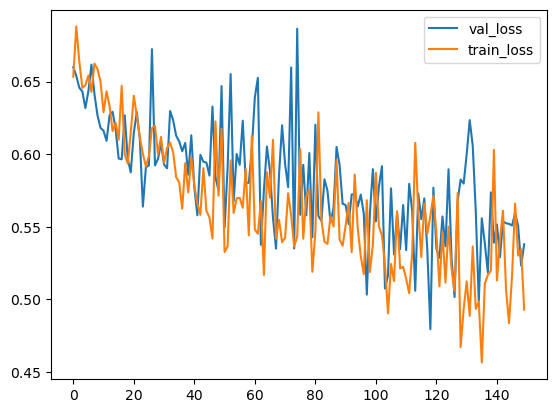} &
          \cellcolor[HTML]{D5DBDB} \includegraphics[height=3.0cm,width=4cm]{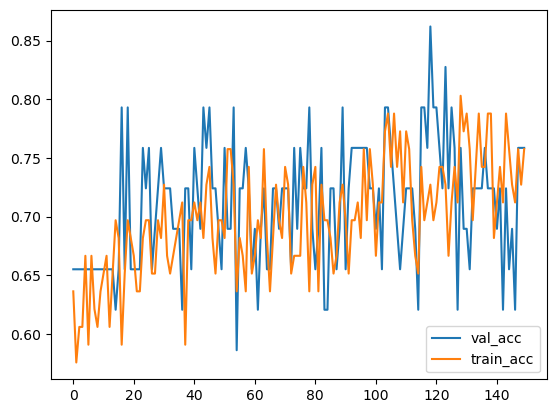} &
          \includegraphics[height=3.0cm,width=4cm]{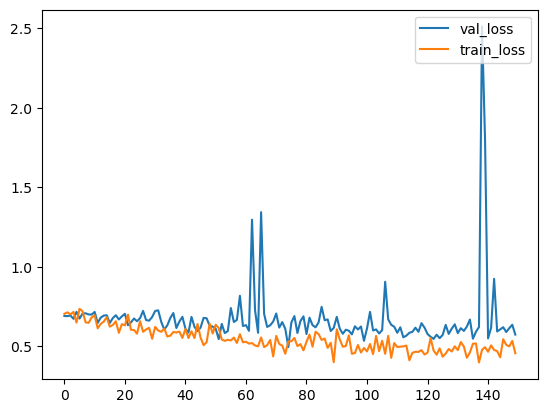} &
          \includegraphics[height=3.0cm,width=4cm]{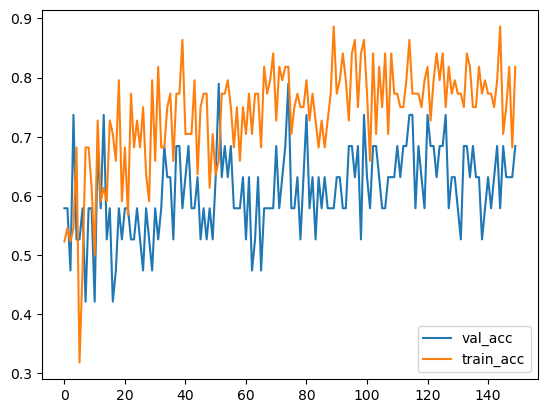}
        \\ \hline
        
        \rotatebox[origin=c]{90}{\centering {MV Profiles}} &
          \includegraphics[height=3.0cm,width=4cm]{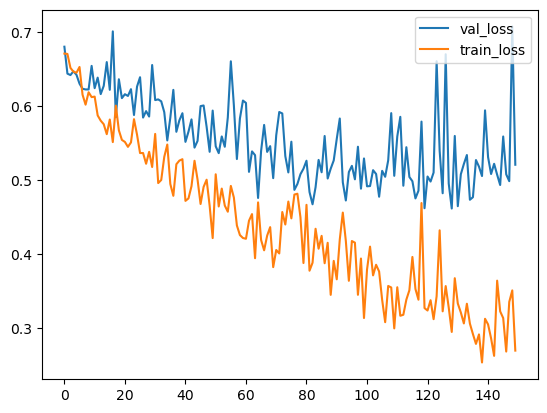} &
          \includegraphics[height=3.0cm,width=4cm]{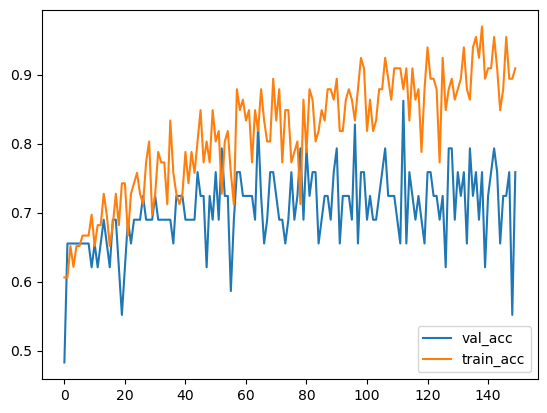} &
          \includegraphics[height=3.0cm,width=4cm]{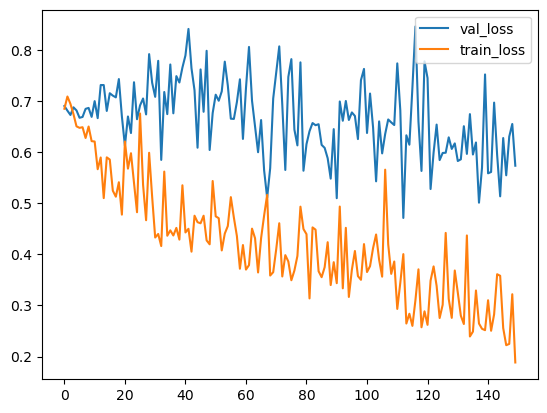} &
          \includegraphics[height=3.0cm,width=4cm]{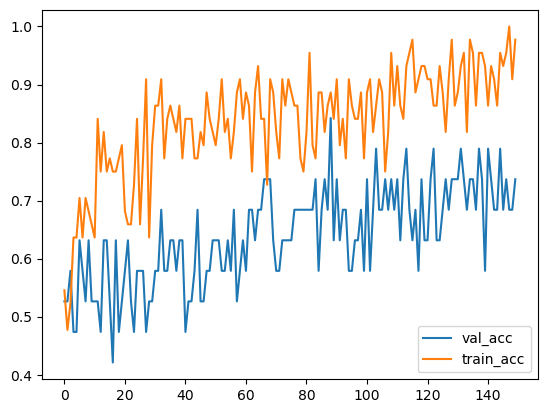}
        \\ \hline
        
        \rotatebox[origin=c]{90}{\centering {MV Full Group}} &
          \cellcolor[HTML]{D5DBDB} \includegraphics[height=3.0cm,width=4cm]{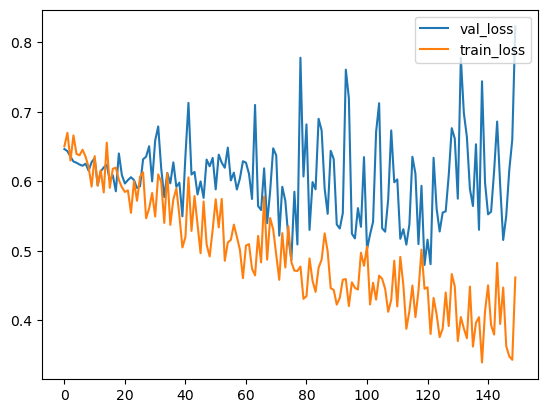} &
          \cellcolor[HTML]{D5DBDB} \includegraphics[height=3.0cm,width=4cm]{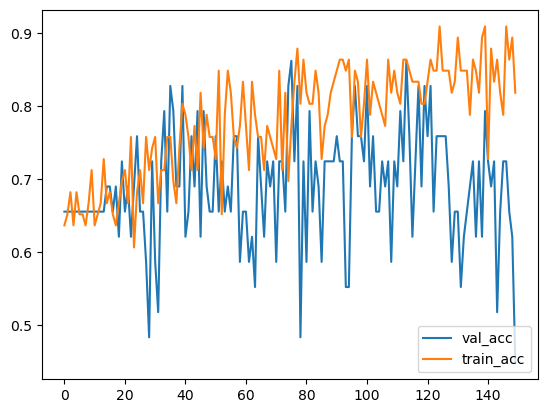} &
          \includegraphics[height=3.0cm,width=4cm]{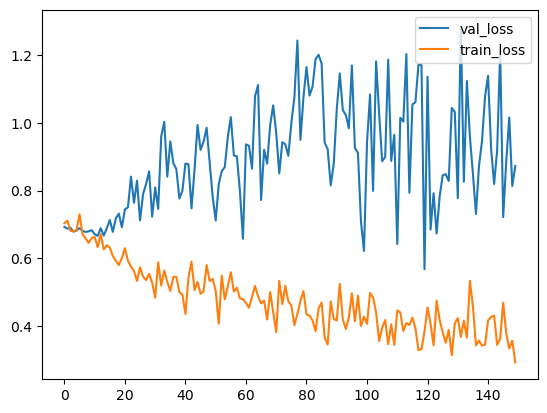} &
          \includegraphics[height=3.0cm,width=4cm]{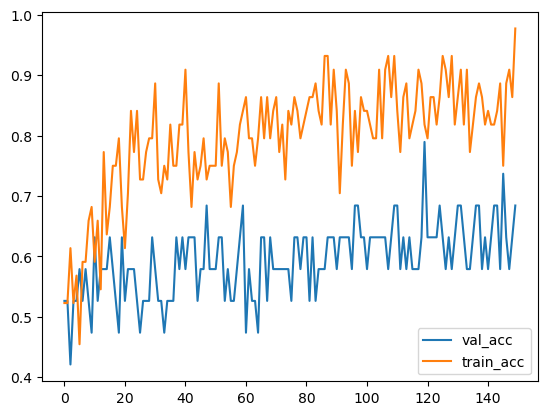}
        \\ \hline
      \end{tabular}
      \end{center}
      \caption{Learning curves (accuracy and loss) of all models. +ND and -ND stand for normal-defective and represent the models' training and testing with or without normal-defective examples. MV stands for a multi-view model and OV stands for a one-view model. Best models' cells are shadowed.}
      \label{table:learning_curve}
      \end{table*}


In cases of a shortage of data, transfer learning is usually a decent solution. However, as mentioned above, an over-complicated model prunes to overfit. Thus, we applied the following methods to the model to overcome the problem: 
\begin{enumerate}
    \item Simplify the model by cutting some of its last layers (Fig. \ref{fig:multi_view_schema} C). The 18 last convolutional layers have been cut, and the network's last layer has been reduced to 128 neurons.
 \item Changing input dimension for 1x244x244 instead of 3x224x224 -- since ResNet has been trained on a colored data set -- differently from our input examples which are in grayscale -- the dimensional change decreases lots of unnecessary weights.
\end{enumerate}
 Overall, there are 1,341,890 learnable parameters instead of 21,799,674 -- about 6\% of the original network.

\begin{figure}[!hb]
    \centering
    \includegraphics[height=5.5cm, trim={4cm 17.7cm 5.4cm 4.1cm},clip]{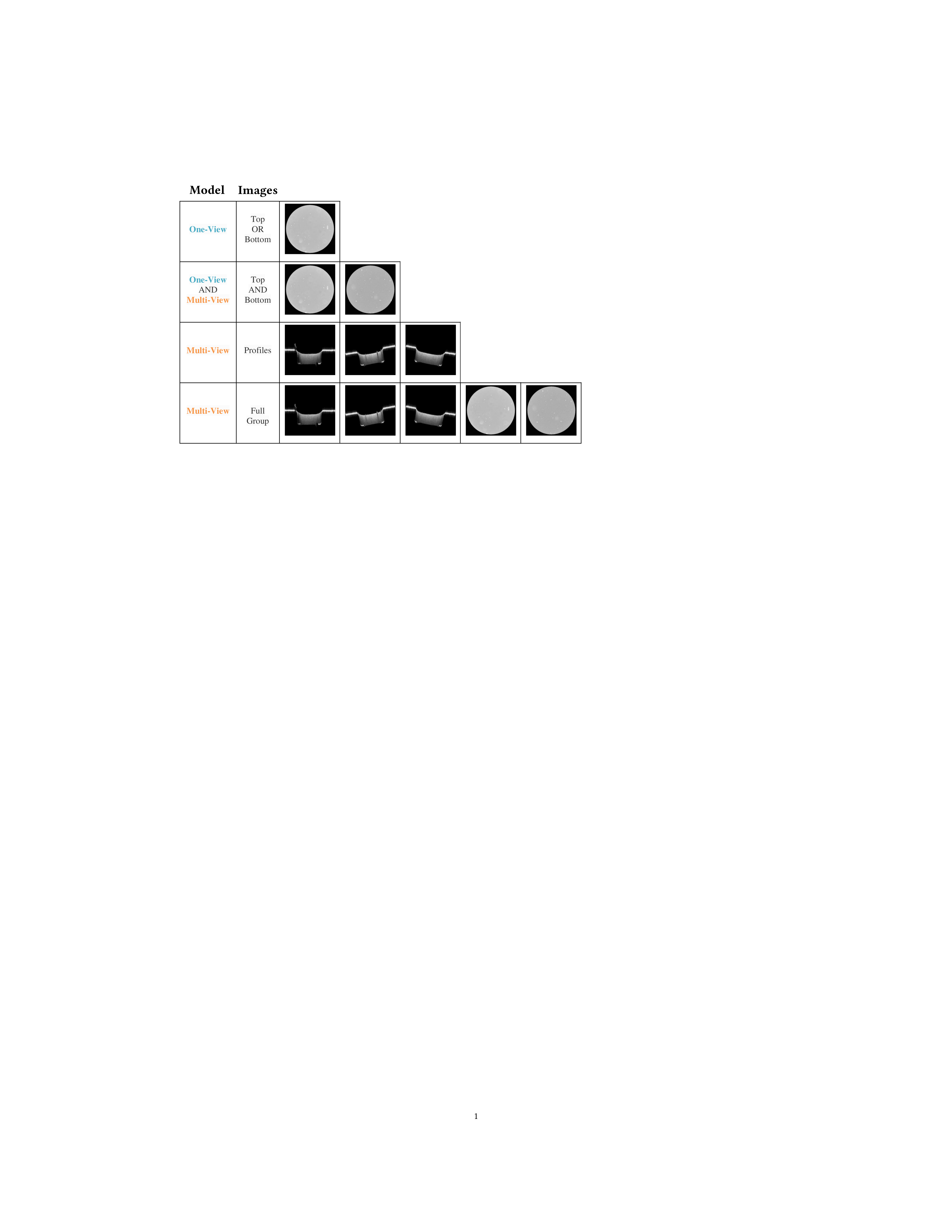}
    \caption{Six learning configurations based on the different types of models and image sets.}
    \label{fig:pyramid}
\end{figure}

Searching for the optimal performance, we survey 6 model configurations (Fig. \ref{fig:pyramid}). Specifically, three one-view models with solely top view, bottom view, and a combination of both; three multi-view models with both top and bottom views; all three profiles; and a full group of five images. Each of the 6 models was trained on the above-stated data set with and without normal-defective examples (marked in Table \ref{table:accuracy} and Table \ref{table:auc} as ND -- normal-defected).

We hereby elaborate on the relevant performance metrics (\ref{performance}), the effect of data additions on the model performances (\ref{added_data}), the yielded experimental results (\ref{results}) and the possibility for explainability (\ref{explainability}).

\subsection{Performance Metrics}\label{performance}
Two metrics have been used for measuring the models' performance. The first is the model's accuracy in the epoch with the minimum loss (Table \ref{table:accuracy}). This measurement indicates the model's performance on the particular test set. Another metric that has been used is the area under curve (AUC) \cite{marzban2004roc} (Table \ref{table:auc}). AUC measures the model's generalization capabilities and, therefore, its performance not only on a certain test set but on many. An additional noteworthy parameter is the loss graph, indicating if convergence and real learning have occurred.

\subsection{Additional Data Effect}\label{added_data}
To understand the impact of the data set size on the resulted model performances, a comparison between two train sets has been performed: The train set mentioned above was compared to itself with subtraction of 20 examples, while the test set remains similar for both sets. This test has been performed to determine unequivocally that there is a need for continued expansion of the data set to achieve better results and that the model truthfully learns from the data and does not memorize it. The addition of 20 training examples has shown an unambiguous improvement in the models' performances -- an improvement of 8\% for every model's accuracy on average and a 16.5\% improvement for every model's AUC on average.

\subsection{Experimental Results}\label{results}
The data set split was a 70:30 train-to-test ratio while maintaining a balanced ratio between classes to prevent the model from learning from skewed data. Overall, there are 66 training examples with 23, 21, and 22 normal, normal-defective, and defective labeling correspondingly and 29 test examples with 10, 10, and 9 normal, normal-defective, and defective labeling correspondingly. The models have run on 32GB Tesla V100 GPU, using Pytorch \cite{paszke2019pytorch}. Accuracy and AUC measures are shown in Table \ref{table:accuracy} and Table \ref{table:auc}, respectively. Loss and accuracy trends are presented in Table \ref{table:learning_curve}. Conclusively, we show that the multi-view model with top-bottom views has the best accuracy -- 86\%, and the multi-view model with the entire group has the best AUC -- 84\%.

\subsection{Model Explainability}\label{explainability}
\begin{figure}[!ht]
\centering
\captionsetup[subfigure]{labelformat=empty}
\begin{subfigure}{.2\textwidth}
\raggedleft
\includegraphics[height=3.5cm]{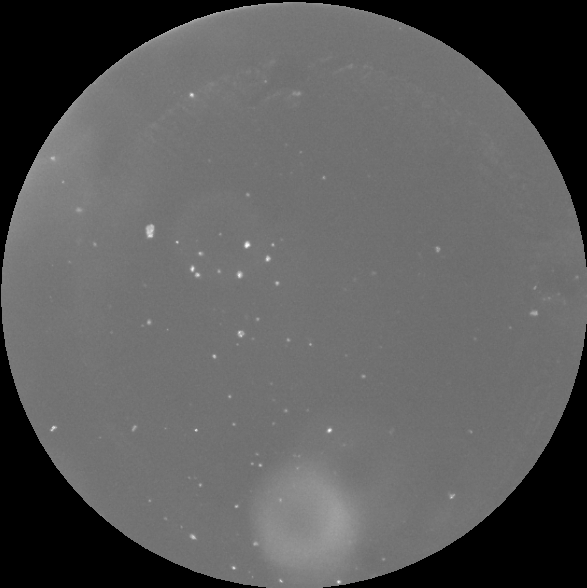}
\caption{Original}
\end{subfigure}
\begin{subfigure}{.2\textwidth}
\raggedright
\includegraphics[height=3.5cm]{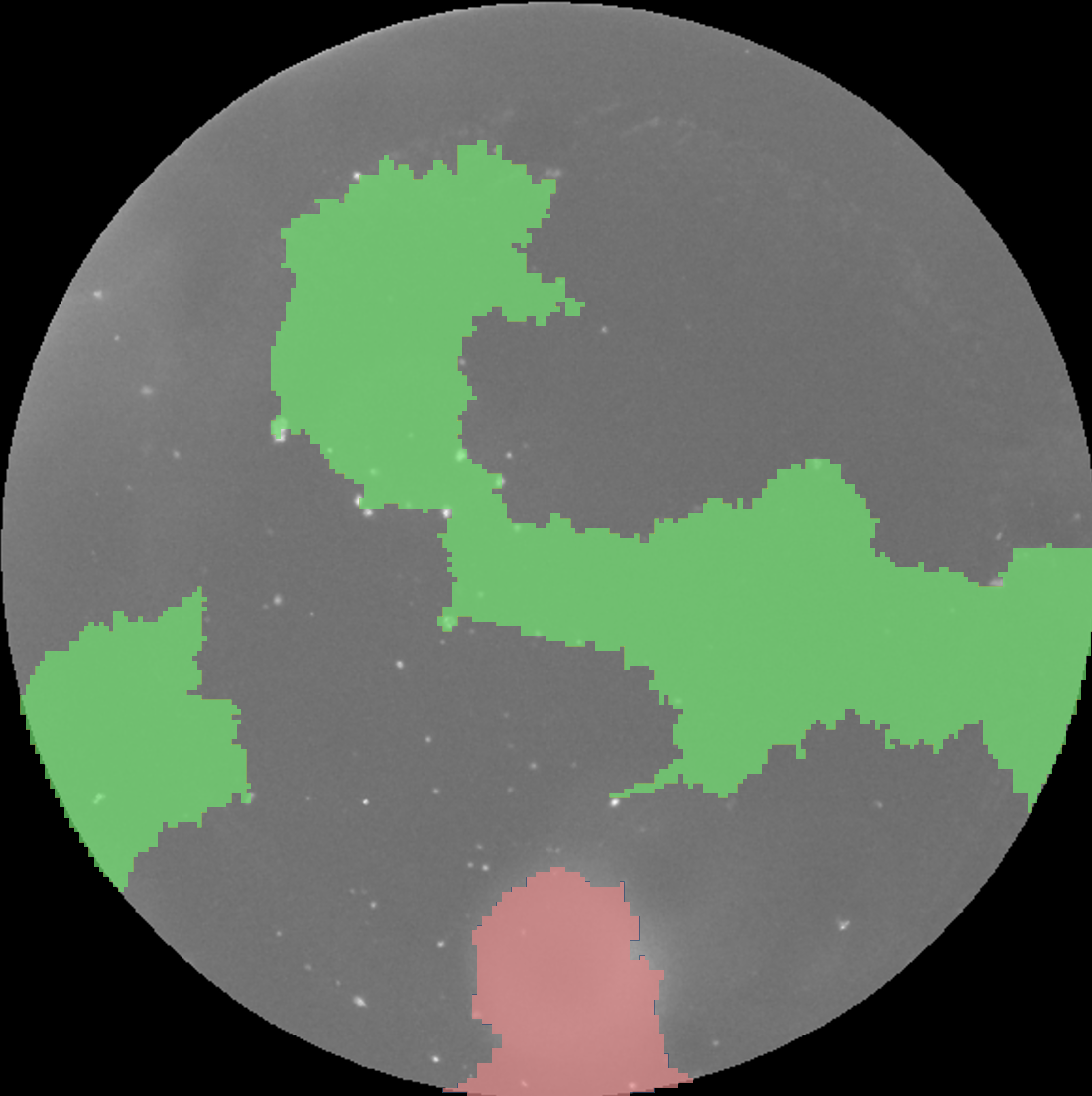}
\caption{LIME's Output}
\end{subfigure}
\caption{Model's explainability using LIME. Contributing areas for normal or defective decision are green and red, respectively.}
\label{fig:lime_example}
\end{figure}

Physics-oriented models generally need to ensure that the outcome predictions are based on actual, relevant features and not on imaginary correlations. Thus, LIME \cite{lime} algorithm was implemented on the model predictions to study the connections between inputs and outputs. Specifically on the one-view top model as it is more intuitive to comprehend. In LIME's output, contributing areas for normal or defective decisions are marked as green or red, respectively. White stain is colored under the red area while the green area is mostly clear -- both truthfully explaining the original rationale (Fig. \ref{fig:lime_example}).


\section{Conclusions and Future Work}

While there is high accuracy for the multi-view model with top-bottom views (86\%), its AUC is relatively low (71\%), meaning the model managed to perform well on the test set, but its generalization capability is incomplete. Nevertheless, the loss graph shows a significant learning process without any overfit. On the other hand, the multi-view model with the entire group (full group) has the best AUC (84\%), and its accuracy is relatively high as well (82\%). However, there is an obvious overfit in the early epochs of the training.

Most of our multi-view models show better results with the normal-defective examples in accuracy and AUC. A possible explanation may lie in the nature of normal-defective examples. Those examples are similar to normal, however, still have tiny defects. The action of labeling those examples as defective may contribute to learning these nuances and improve the classification.

For future work, we can name the apparent need for data set expansion while maintaining class balance. Also, further investigation of different pre-trained models for better transfer learning can be helpful. For example, the Inception pre-trained model \cite{szegedy2017inception}, which has been trained on vast x-rays data set, may be more fitted for extracting features from the profiles. Finally, we suggest examining the possibility of concatenating the views' feature maps which preserves and assumes order, similarly to the given problem, instead of max-pooling them.

\section*{Acknowledgment}
This work was supported by the Pazy foundation and the Lynn and William Frankel Center for Computer Science. Computational support was provided by the NegevHPC project~\cite{negevhpc}. The authors would like to thank Galit Bar\IEEEauthorrefmark{5}, Guy Lazovski\IEEEauthorrefmark{5} and Muriel Tzadka\IEEEauthorrefmark{1} for foam samples preparation.

\bibliographystyle{IEEEtran}
\bibliography{IEEEabrv,sample-base.bib}

\begin{thebibliography}{10}
\providecommand{\url}[1]{#1}
\csname url@samestyle\endcsname
\providecommand{\newblock}{\relax}
\providecommand{\bibinfo}[2]{#2}
\providecommand{\BIBentrySTDinterwordspacing}{\spaceskip=0pt\relax}
\providecommand{\BIBentryALTinterwordstretchfactor}{4}
\providecommand{\BIBentryALTinterwordspacing}{\spaceskip=\fontdimen2\font plus
\BIBentryALTinterwordstretchfactor\fontdimen3\font minus
  \fontdimen4\font\relax}
\providecommand{\BIBforeignlanguage}[2]{{%
\expandafter\ifx\csname l@#1\endcsname\relax
\typeout{** WARNING: IEEEtran.bst: No hyphenation pattern has been}%
\typeout{** loaded for the language `#1'. Using the pattern for}%
\typeout{** the default language instead.}%
\else
\language=\csname l@#1\endcsname
\fi
#2}}
\providecommand{\BIBdecl}{\relax}
\BIBdecl

\bibitem{ref1}
K.~Mussack \emph{et~al.}, ``Simulating hohlraum dynamics and radiation flow for
  pleiades experiments on nif,'' in \emph{APS Division of Plasma Physics
  Meeting Abstracts}, vol.~53, 2011, pp. GP9--124.

\bibitem{ref2}
M.~Baer, ``A numerical study of shock wave reflections on low density foam,''
  \emph{Shock Waves}, vol.~2, no.~2, pp. 121--124, 1992.

\bibitem{ref3}
H.~M. Johns \emph{et~al.}, ``A temperature profile diagnostic for radiation
  waves on omega-60,'' \emph{High Energy Density Physics}, vol.~39, p. 100939,
  2021.

\bibitem{ref4}
Y.~Wu \emph{et~al.}, ``Dynamics of laser produced plasma from foam targets for
  future nanolithography devices and x-ray sources,'' \emph{Scientific
  Reports}, vol.~11, no.~1, pp. 1--14, 2021.

\bibitem{harel2020complete}
R.~Harel \emph{et~al.}, ``Complete deep computer-vision methodology for
  investigating hydrodynamic instabilities,'' in \emph{International Conference
  on High Performance Computing}.\hskip 1em plus 0.5em minus 0.4em\relax
  Springer, 2020, pp. 61--80.

\bibitem{ref7}
O.~Figovsky \emph{et~al.}, \emph{Green nanotechnology}.\hskip 1em plus 0.5em
  minus 0.4em\relax Jenny Stanford Publishing, 2017.

\bibitem{ref8}
D.~W. Schaefer \emph{et~al.}, ``Structure of random porous materials: silica
  aerogel,'' \emph{Physical review letters}, vol.~56, no.~20, p. 2199, 1986.

\bibitem{ref9}
G.~Wei \emph{et~al.}, ``Thermal conductivities study on silica aerogel and its
  composite insulation materials,'' \emph{International Journal of Heat and
  Mass Transfer}, vol.~54, no. 11-12, pp. 2355--2366, 2011.

\bibitem{ref10}
M.~Reim \emph{et~al.}, ``S., m. arduini-schuster, hp ebert, j. fricke, j,''
  \emph{Silica aerogel granulate material for thermal insulation and
  daylighting, Solar Energy}, vol.~79, pp. 131--139, 2005.

\bibitem{ref11}
B.~Yoldas \emph{et~al.}, ``Chemical engineering of aerogel morphology formed
  under nonsupercritical conditions for thermal insulation,'' \emph{Chemistry
  of materials}, vol.~12, no.~8, pp. 2475--2484, 2000.

\bibitem{ref12}
E.~Zhang \emph{et~al.}, ``Insulating and robust ceramic nanorod aerogels with
  high-temperature resistance over 1400 c,'' \emph{ACS Applied Materials \&
  Interfaces}, vol.~13, no.~17, pp. 20\,548--20\,558, 2021.

\bibitem{ref13}
S.~Takeshita \emph{et~al.}, ``Structural and acoustic properties of transparent
  chitosan aerogel,'' \emph{Materials Letters}, vol. 254, pp. 258--261, 2019.

\bibitem{ref14}
K.~Wu \emph{et~al.}, ``Lightweight and flexible phenolic aerogels with
  three-dimensional foam reinforcement for acoustic and thermal insulation,''
  \emph{Industrial \& Engineering Chemistry Research}, vol.~60, no.~3, pp.
  1241--1249, 2021.

\bibitem{ref15}
Y.~Gu \emph{et~al.}, ``\BIBforeignlanguage{English (US)}{Electronic structure
  tuning in ni3fen/r-go aerogel toward bifunctional electrocatalyst for overall
  water splitting},'' \emph{\BIBforeignlanguage{English (US)}{ACS Nano}},
  vol.~12, no.~1, pp. 245--253, Jan. 2018.

\bibitem{ref16}
M.~A. Worsley \emph{et~al.}, ``Synthesis of graphene aerogel with high
  electrical conductivity,'' \emph{Journal of the American Chemical Society},
  vol. 132, no.~40, pp. 14\,067--14\,069, 2010.

\bibitem{ref17}
Peles-Strahl \emph{et~al.}, ``Bipyridine modified conjugated carbon aerogels as
  a platform for the electrocatalysis of oxygen reduction reaction,''
  \emph{Advanced Functional Materials}, vol.~31, no.~26, p. 2100163, 2021.

\bibitem{ref18}
N.~U. Wetter \emph{et~al.}, ``Dynamic random lasing in silica aerogel doped
  with rhodamine 6g,'' \emph{RSC advances}, vol.~8, no.~52, pp.
  29\,678--29\,685, 2018.

\bibitem{ref19}
M.~J. Burchell \emph{et~al.}, ``Characteristics of cometary dust tracks in
  stardust aerogel and laboratory calibrations,'' \emph{Meteoritics \&
  Planetary Science}, vol.~43, no. 1-2, pp. 23--40, 2008.

\bibitem{ref20}
N.~Ganonyan \emph{et~al.}, ``Entrapment of enzymes in silica aerogels,''
  \emph{Materials Today}, vol.~33, pp. 24--35, 2020.

\bibitem{ref21}
G.~Nir \emph{et~al.}, ``Gradual hydrophobization of silica aerogel for
  controlled drug release,'' \emph{RSC advances}, vol.~11, no.~14, pp.
  7824--7838, 2021.

\bibitem{ref22}
Garc{\'\i}a-Gonz{\'a}lez \emph{et~al.}, ``Polysaccharide-based aerogel
  microspheres for oral drug delivery,'' \emph{Carbohydrate polymers}, vol.
  117, pp. 797--806, 2015.

\bibitem{ref23}
J.~Zhao \emph{et~al.}, ``Polyethylenimine-grafted cellulose nanofibril aerogels
  as versatile vehicles for drug delivery,'' \emph{ACS applied materials \&
  interfaces}, vol.~7, no.~4, pp. 2607--2615, 2015.

\bibitem{ref24}
N.~Ganonyan \emph{et~al.}, ``Ultralight monolithic magnetite aerogel,''
  \emph{Applied Materials Today}, vol.~22, p. 100955, 2021.

\bibitem{ref25}
D.~Asner \emph{et~al.}, ``Experimental study of aerogel cherenkov detectors for
  particle identification,'' \emph{Nuclear Instruments and Methods in Physics
  Research Section A: Accelerators, Spectrometers, Detectors and Associated
  Equipment}, vol. 374, no.~3, pp. 286--292, 1996.

\bibitem{ref26}
T.~Sumiyoshi \emph{et~al.}, ``Silica aerogel cherenkov counter for the kek
  b-factory experiment,'' \emph{Nuclear Instruments and Methods in Physics
  Research Section A: Accelerators, Spectrometers, Detectors and Associated
  Equipment}, vol. 433, no. 1-2, pp. 385--391, 1999.

\bibitem{ref27}
C.~J. Brinker \emph{et~al.}, \emph{Sol-gel science: the physics and chemistry
  of sol-gel processing}.\hskip 1em plus 0.5em minus 0.4em\relax Academic
  press, 2013.

\bibitem{ref28}
R.~Gvishi, ``Monolith sol-gel materials,'' \emph{chapter in [The Sol-Gel
  Handbook], Eds: Levy D. and Zayat M., in publication by Wiley-VCH}, 2014.

\bibitem{ref29}
I.~Smirnova \emph{et~al.}, ``Aerogel production: Current status, research
  directions, and future opportunities,'' \emph{The Journal of Supercritical
  Fluids}, vol. 134, pp. 228--233, 2018.

\bibitem{ref30}
G.~Lazovski \emph{et~al.}, ``A simple method for preparation of silica aerogels
  doped with monodispersed nanoparticles in homogeneous concentration,''
  \emph{The Journal of Supercritical Fluids}, vol. 159, p. 104496, 2020.

\bibitem{ref31}
L.~Guy \emph{et~al.}, ``A procedure to synthesize silica aerogels in a wide
  range of densities by a single-step base catalyzed recipe,'' \emph{Journal of
  Porous Materials}, vol.~28, no.~4, pp. 1227--1236, 2021.

\bibitem{ref32}
A.~Sedova \emph{et~al.}, ``Reinforcing silica aerogels with tungsten disulfide
  nanotubes,'' \emph{The Journal of Supercritical Fluids}, vol. 106, pp. 9--15,
  2015.

\bibitem{ref33}
R.~Haj-Ali \emph{et~al.}, ``Mechanical characterization of aerogel materials
  with digital image correlation,'' \emph{Microporous and Mesoporous
  Materials}, vol. 226, pp. 44--52, 2016.

\bibitem{ref34}
A.~Sedova \emph{et~al.}, ``Silica aerogels as hosting matrices for ws2
  nanotubes and their optical characterization,'' \emph{Journal of materials
  science}, vol.~55, no.~18, pp. 7612--7623, 2020.

\bibitem{rusanovsky2022end}
M.~Rusanovsky \emph{et~al.}, ``An end-to-end computer vision methodology for
  quantitative metallography,'' \emph{Scientific Reports}, vol.~12, no.~1, pp.
  1--27, 2022.

\bibitem{rusanovsky2020mlography}
R.~Matan \emph{et~al.}, ``Mlography: An automated quantitative metallography
  model for impurities anomaly detection using novel data mining and deep
  learning approach,'' \emph{arXiv preprint arXiv:2003.04226}, 2020.

\bibitem{xu2013survey}
C.~Xu \emph{et~al.}, ``A survey on multi-view learning,'' \emph{arXiv preprint
  arXiv:1304.5634}, 2013.

\bibitem{sun2013survey}
S.~Sun, ``A survey of multi-view machine learning,'' \emph{Neural computing and
  applications}, vol.~23, no.~7, pp. 2031--2038, 2013.

\bibitem{zhao2017multi}
J.~Zhao \emph{et~al.}, ``Multi-view learning overview: Recent progress and new
  challenges,'' \emph{Information Fusion}, vol.~38, pp. 43--54, 2017.

\bibitem{seeland2021multi}
M.~Seeland \emph{et~al.}, ``Multi-view classification with convolutional neural
  networks,'' \emph{Plos one}, vol.~16, no.~1, 2021.

\bibitem{nie2017multi}
F.~Nie \emph{et~al.}, ``Multi-view clustering and semi-supervised
  classification with adaptive neighbours,'' in \emph{Thirty-first AAAI
  conference on artificial intelligence}, 2017.

\bibitem{houthuys2018multi}
L.~Houthuys \emph{et~al.}, ``Multi-view least squares support vector machines
  classification,'' \emph{Neurocomputing}, vol. 282, pp. 78--88, 2018.

\bibitem{kan2016multi}
M.~Kan \emph{et~al.}, ``Multi-view deep network for cross-view
  classification,'' in \emph{Proceedings of the IEEE Conference on Computer
  Vision and Pattern Recognition}, 2016, pp. 4847--4855.

\bibitem{kumar2014detailed}
G.~Kumar \emph{et~al.}, ``A detailed review of feature extraction in image
  processing systems,'' in \emph{2014 Fourth international conference on
  advanced computing \& communication technologies}.\hskip 1em plus 0.5em minus
  0.4em\relax IEEE, 2014, pp. 5--12.

\bibitem{juan2009comparison}
L.~Juan \emph{et~al.}, ``A comparison of sift, pca-sift and surf,''
  \emph{International Journal of Image Processing (IJIP)}, vol.~3, no.~4, pp.
  143--152, 2009.

\bibitem{hartigan1979algorithm}
J.~A. Hartigan \emph{et~al.}, ``Algorithm as 136: A k-means clustering
  algorithm,'' \emph{Journal of the royal statistical society. series c
  (applied statistics)}, vol.~28, no.~1, pp. 100--108, 1979.

\bibitem{multiweb}
S.~Tafasca, ``Multi-view image classification: From logistic regression to
  multi-view convolutional neural networks (mvcnn),''
  \url{https://towardsdatascience.com/multi-view-image-classification-427c69720f30},
  December 24, 2019, [Online].

\bibitem{khan2018guide}
S.~Khan \emph{et~al.}, ``A guide to convolutional neural networks for computer
  vision,'' \emph{Synthesis lectures on computer vision}, vol.~8, no.~1, pp.
  1--207, 2018.

\bibitem{su2015multi}
H.~Su \emph{et~al.}, ``Multi-view convolutional neural networks for 3d shape
  recognition,'' in \emph{Proceedings of the IEEE international conference on
  computer vision}, 2015, pp. 945--953.

\bibitem{weiss2016survey}
K.~Weiss \emph{et~al.}, ``A survey of transfer learning,'' \emph{Journal of Big
  data}, vol.~3, no.~1, pp. 1--40, 2016.

\bibitem{he2016deep}
K.~He \emph{et~al.}, ``Deep residual learning for image recognition,'' in
  \emph{Proceedings of the IEEE conference on computer vision and pattern
  recognition}, 2016, pp. 770--778.

\bibitem{russakovsky2015imagenet}
O.~Russakovsky \emph{et~al.}, ``Imagenet large scale visual recognition
  challenge,'' \emph{International journal of computer vision}, vol. 115,
  no.~3, pp. 211--252, 2015.

\bibitem{kadam2018review}
S.~Kadam \emph{et~al.}, ``Review and analysis of zero, one and few shot
  learning approaches,'' in \emph{International Conference on Intelligent
  Systems Design and Applications}.\hskip 1em plus 0.5em minus 0.4em\relax
  Springer, 2018, pp. 100--112.

\bibitem{canny1986computational}
J.~Canny, ``A computational approach to edge detection,'' \emph{IEEE
  Transactions on pattern analysis and machine intelligence}, no.~6, pp.
  679--698, 1986.

\bibitem{marzban2004roc}
C.~Marzban, ``The roc curve and the area under it as performance measures,''
  \emph{Weather and Forecasting}, vol.~19, no.~6, pp. 1106--1114, 2004.

\bibitem{paszke2019pytorch}
A.~Paszke \emph{et~al.}, ``Pytorch: An imperative style, high-performance deep
  learning library,'' \emph{Advances in neural information processing systems},
  vol.~32, 2019.

\bibitem{lime}
M.~T. Ribeiro \emph{et~al.}, ``"why should i trust you?": Explaining the
  predictions of any classifier,'' in \emph{Proceedings of the 22nd {ACM}
  {SIGKDD} International Conference on Knowledge Discovery and Data Mining, San
  Francisco, CA, USA,}, August 13-17, 2016, pp. 1135--1144.

\bibitem{szegedy2017inception}
C.~Szegedy \emph{et~al.}, ``Inception-v4, inception-resnet and the impact of
  residual connections on learning,'' in \emph{Thirty-first AAAI conference on
  artificial intelligence}, 2017.

\bibitem{negevhpc}
R.~I. Park, ``{NegevHPC Project},'' \url{https://www.negevhpc.com}, 2019,
  [Online].

\end{thebibliography}

\end{document}